\documentclass[conference]{IEEEtran}
\IEEEoverridecommandlockouts
\usepackage{cite}
\usepackage[table]{xcolor}
\usepackage{amsmath,amssymb,amsfonts}
\usepackage{algorithmic}
\usepackage{graphicx}
\usepackage{amsthm}
\newtheorem{definition}{Problem}
\usepackage{tabularx}
\usepackage{subcaption}
\usepackage{booktabs}
\usepackage[most]{tcolorbox}
\usepackage{textcomp}
\usepackage{url}
\usepackage{xcolor}
\usepackage{textcomp}
\usepackage{siunitx}
\usepackage[gen]{eurosym}
\def\BibTeX{{\rm B\kern-.05em{\sc i\kern-.025em b}\kern-.08em
    T\kern-.1667em\lower.7ex\hbox{E}\kern-.125emX}}
\begin{document}

\title{\textit{From Benchmarks to Production:} \\ Transferring Time Series Anomaly Detection Methods for Electricity Production Monitoring}


\author{\IEEEauthorblockN{Nicolas Vautier$^1$, Paul Caron$^2$, Nardi Xhepi$^2$, Félicie Bizeul$^1$, Manel Boumghar$^1$, Christophe Degouy$^2$, Paul Boniol$^3$}
\IEEEauthorblockA{\textit{$^1$ EDF Lab Paris Saclay, $^2$ EDF, DOAAT, $^3$ Inria, Ecole Normale Supérieure (PSL), CNRS} \\
$^{1,2}$ firstname.lastname@edf.fr, $^3$ paul.boniol@inria.fr}
}





\maketitle


\begin{abstract}

Accurate forecasting of electricity production is essential for maintaining the operational efficiency and strategic planning of energy utilities. In industrial settings, such forecasts are generated daily to ensure supply–demand balance and optimal management of production assets. However, the increasing complexity of modern power systems and data flows poses significant challenges for ensuring the reliability and consistency of these forecasts. This paper addresses the problem of anomaly detection in short-term production forecasts at EDF, formulated as identifying atypical intra-day patterns that may signal data quality issues or operational irregularities. We introduce \textsc{TAMIS}, a scalable and interpretable system that analyzes daily production time series to automatically detect anomalous days based on deviations from historical patterns learned from past data. Designed for human-in-the-loop workflows, \textsc{TAMIS} surfaces top-ranked anomalies through an automated daily newsletter, enabling efficient expert review and continuous monitoring. An extensive experimental evaluation on real-world industrial data demonstrates that \textsc{TAMIS} achieves the best accuracy–efficiency trade-off compared to baseline methods. To foster further research and reproducibility, we publicly release the anonymized application datasets used in our study.
  
\end{abstract}

\maketitle

\section{Introduction}

Time series data play a central role in a wide range of industrial applications~\cite{Palpanas2019,uehara2002extraction,bach2017flexible}, from predictive maintenance and fault detection~\cite{Boniol_Meftah_Remy_Didier_Palpanas_2023} to resource optimization and demand forecasting~\cite{camal}. A common challenge across these domains is the reliable detection of unusual patterns or behaviors that deviate from historical data (most commonly called \emph{anomaly} or \emph{outlier} in the literature~\cite{Series2GraphPaper,Schmidl22,DBLP:journals/pvldb/BoniolPPF21}). Anomaly detection in time series is particularly critical in environments where real-time decisions must be made based on automatically generated forecasts or measurements. Despite significant progress in machine learning and statistical methods for time series analysis~\cite{boniol2024divetimeseriesanomalydetection, 10.1145/3711896.3736565}, the practical deployment of anomaly detection systems in industry remains difficult due to specific constraints that often arise in operational settings.

This paper is motivated by the real-world context of energy production planning at EDF, one of the largest electricity producers in Europe. In this setting, large volumes of heterogeneous time series are generated daily to forecast the operation of power plants and storage systems over a 24-hour horizon. Each forecast is represented as a fixed-length vector capturing 48 half-hourly values for the upcoming day. These forecasts are critical inputs to balancing supply and demand, planning energy dispatch, and engaging with energy markets. Ensuring their reliability is thus essential. However, due to the scale and complexity of the data, it is not feasible for domain experts to manually review all forecasts, which necessitates an automated, trustworthy anomaly detection solution. More precisely, our industrial constraints are as follows:
First, anomaly detection must be \textbf{online}, in the sense that it evaluates whether the most recent day-ahead forecast is abnormal based on historical data up to that point. Second, the solution must be \textbf{interpretable and human-centered}, as detection results are analyzed daily by energy experts who require concise, actionable explanations. Third, the method must be \textbf{scalable}, capable of handling large volumes of time series, with low latency, and limited hardware resources. These constraints rule out many existing approaches that rely on deep learning models, exhaustive offline training.

\begin{figure}
    \centering
    \includegraphics[width=\linewidth]{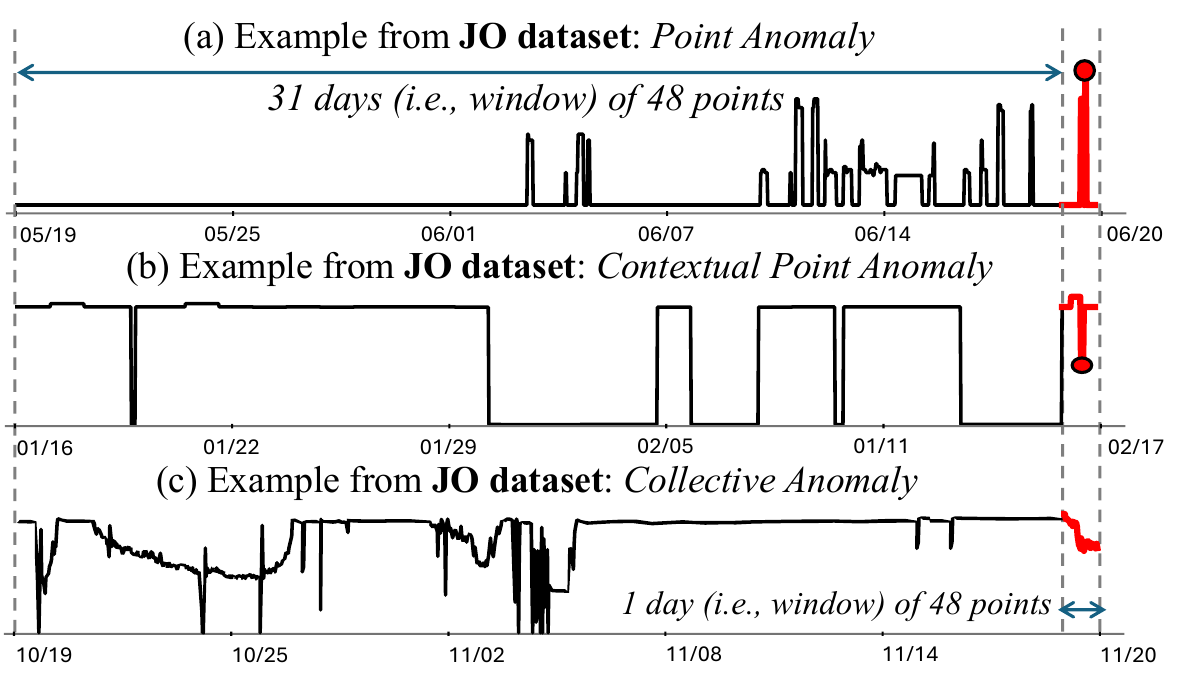}
    \caption{Examples of time series in \textbf{our proposed \textsc{JO} dataset} (anomalies are highlighted in red).}
    \label{fig:introfig}
\end{figure}

To address these challenges, we propose \textsc{TAMIS}, a lightweight and interpretable anomaly detection system designed for production-grade deployment. Each daily forecast is scored for abnormality using a feature-based approach. The results are automatically ranked and integrated into a daily report. Only the most abnormal cases are returned, allowing experts to review them and make decisions accordingly. 

Beyond proposing a practical solution, this work also provides insights into the broader journey of moving from academic research on time series anomaly detection to an operational system. We discuss how methodological advances translate into real-world applicability, and how constraints such as hardware efficiency, scalability, and expert usability shape the design of a deployable framework. In this sense, the paper not only introduces a new system, but also illustrates the path from research concepts to an industry-ready solution.

Finally, we evaluate our system on a large, real-world dataset collected from EDF's internal production forecasts over multiple years. This dataset is diverse in both physical nature and statistical behavior, reflecting the complexity of our use case. We perform a rigorous experimental study to assess the scalability and detection accuracy of different algorithmic design within our framework. Furthermore, we release a curated and anonymized version of this dataset as a public benchmark to encourage future research in time series anomaly detection.
Overall, our contributions are as follows:

\begin{itemize}
    \item \textbf{Novel Industrial Architecture:} We analyze the requirements of time series anomaly detection in our industrial setting and propose a novel architecture tailored to these practical operational constraints (\textbf{Sec.~\ref{sec:IndusandProb}}).
    \item \textbf{Designed-for-Purpose Components:} We introduce a lightweight detector, \textbf{ASHES}, and \textbf{$\text{TAMIS}_{\mathcal{F}}$}, a feature set that forms a core component of our framework (\textbf{Sec.~\ref{sec:ProposedApp}}).
    \item \textbf{An End-to-end Pipeline:} We present \textbf{TAMIS}, a novel end-to-end anomaly detection framework for electrical production time series that balances interpretability, accuracy, and scalability (\textbf{Sec.~\ref{sec:ProposedApp}}).
    \item \textbf{Open Data:} We release three open-access datasets of real, labeled time series, which are among the first of their kind in the energy sector (cf. \textbf{Sec.~\ref{sec:Datasets}}).
    \item \textbf{Extensive Evaluation:} We conduct an extensive experimental evaluation demonstrating that TAMIS outperforms state-of-the-art baselines while maintaining a superior accuracy-efficiency trade-off (\textbf{Sec.~\ref{sec:expEval}}).
\end{itemize}

\section{Industrial context and problem formulation} 
\label{sec:IndusandProb}

\begin{table}[tb]
\setlength{\tabcolsep}{8pt}
\begin{center}
\resizebox{\columnwidth}{!}{%
\begin{tabular}{ll}
\toprule
\textbf{Symbol} & \textbf{Description} \\
\midrule

$\Delta \in \mathbb{N}$ & Sampling interval ($30$ minutes). \\

$M \in \mathbb{N}$ & Number of samples per day ($M=48$). \\

$\boldsymbol{C}_{d} \in \mathbb{R}^{M}$ & Daily window (24-hour profile) for day $d$ \\

$c_m \in \mathbb{R}$ & $m$-th value of $\boldsymbol{C}_{d}$ \\

$\boldsymbol{F}_{d} \in \mathbb{R}^{i}$ & Feature set for day $d$ \\

$\boldsymbol{f}_{j} \in \mathbb{R}$ & $j$-th value of $\boldsymbol{F}_{d}$ \\

$\boldsymbol{T} \in \mathbb{R}^{n \times M}$ & Time series of $n$ consecutive daily windows \\

${H_j} \in \mathbb{R}^{n}$ & Historical values for the $j$-th feature, for a given time series \\

$n \in \mathbb{N}$ & Number of days (i.e., windows) in the time series $\boldsymbol{T}$. \\

$i \in \mathbb{N}$ & Number of features $\boldsymbol{T}$. \\

$\mathcal{A}$ & Anomaly detection function. \\

$D$ & A unique anomaly detection model (i.e., detector). \\

$\mathcal{M}$ & An ensembling or selection method. \\

$S \in [0,1]$ & Anomaly score. \\

\bottomrule
\end{tabular}
}
\captionsetup{justification=centering}
\caption{Summary of main notations used in this paper.}
\vspace{-0.5cm}
\label{tab:notation}
\end{center}
\end{table}

EDF, France’s leading electricity producer, is committed to building a carbon-neutral energy future through a mix of production sources. A critical enabler of this mission is the ability to continuously balance energy supply and demand, a task that relies heavily on accurate forecasts of both production and consumption. Maintaining this balance in the face of dynamic operational conditions and unforeseen events is essential not only for risk mitigation but also for value optimization.

Time series data lie at the heart of this operational ecosystem. These series capture a wide variety of measurements, including electrical power output, reservoir levels, flow rates, turbine states, production targets, and derived variables such as adjustment margins. These series differ in scale, dynamics, semantics, and physical meaning, making manual inspection infeasible. Each day, large volumes of new data are ingested and logged, underscoring the need for \textbf{automated, scalable monitoring tools} to detect and flag abnormal behavior. 

Figure~\ref{fig:introfig} illustrates examples of time series and anomalies of interest in our study. Overall, time series from EDF datasets contains heterogeneous types of time series and anomalies. Figure~\ref{fig:introfig}(a) illustrates a \textit{global point} anomaly (i.e., a single value that deviates from the global distribution of values within the time series), Figure~\ref{fig:introfig}(b) depicts a \textit{contextual point} anomaly (i.e., a point that deviates from the distribution of values within a given window of the time series), and Figure~\ref{fig:introfig}(c) shows a \textit{Collective} anomaly (i.e., abnormal sequence of values that, if taken independently, would be considered normal).

A key aspect of this work is the operationalization of the detection within a \textit{human-in-the-loop} framework. To bridge the gap between model output and practical action, a \textit{mailing list} is implemented. Each day, the system automatically scores and ranks all potential anomalies. Subsequently, a summary report is distributed to a list of experts. This report is intentionally concise, presenting only the highest-ranked anomalies to mitigate alert fatigue and focus expert analysis on the most pressing issues. Each entry in the report is intended to include key contextual information, such as the anomaly score and timestamp, enabling experts to efficiently triage, validate, and prioritize potential anomalies for in-depth investigation.

\subsection{Problem Definition}
\label{sec:Probdef}

\begin{figure*}
    \centering
    \includegraphics[width=\linewidth]{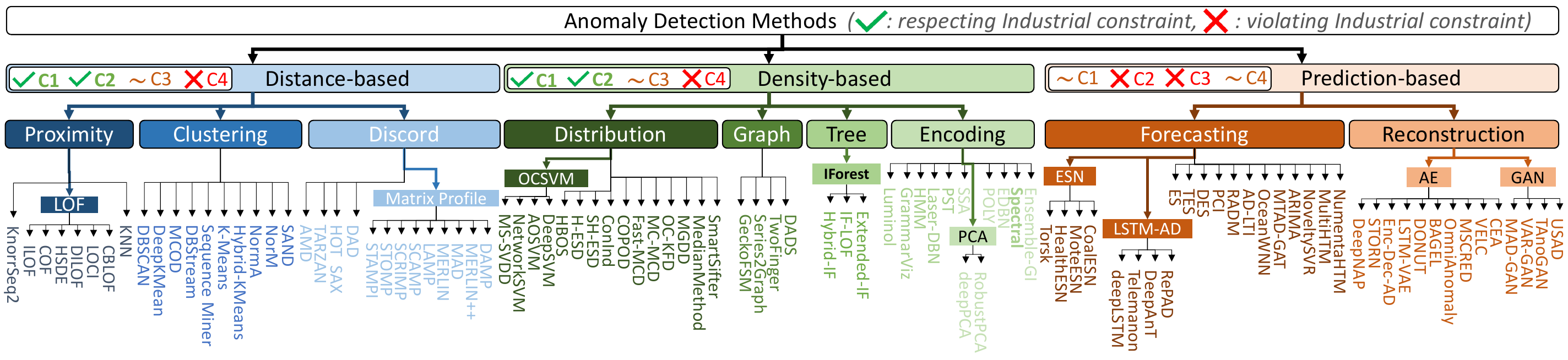}
    \caption{Raw-based Time series anomaly detection methods in the literature~\cite{boniol2024divetimeseriesanomalydetection}}
    \vspace{-0.3cm}
    \label{fig:taxonomy}
\end{figure*}

As mentioned in the previous section, our objective is to identify anomalies in time series, a well-studied problem in the literature for decades~\cite{boniol2024divetimeseriesanomalydetection}.
Anomaly detection in this context is framed as an online, last forecasted window detection task, where the objective is to assess whether the most recently observed forecasted value in a series deviates significantly from historical behavior. Unlike retrospective analyses (the most common tackled problem in the literature~\cite{boniol2024divetimeseriesanomalydetection}), the goal is not to detect previously missed events but to identify anomalies as they emerge, thus enabling timely interventions. This real-time requirement aligns with EDF’s operational needs, where undetected anomalies in key indicators could cascade into significant planning errors or missed opportunities.

We call a \textit{window} the 24-hour profile of a variable \(c\) sampled every \(\Delta=30\) minutes.
Formally, a window $\boldsymbol{C}_{d}$ for a given day $d$ is defined as $\boldsymbol{C}_{d}
= \bigl(c_{1},\ldots,c_{M}\bigr) \in \mathbb{R}^{M}$,
where $M=48$ and 
\(c_{m}\) denotes the measurements for the \(m\)-th half-hour interval of day \(d\). 
Therefore, a time series $T$ is an ordered set of windows defined as follows:

\[
\boldsymbol{T}
= \bigl(\boldsymbol{C}_{d_1},\ldots,\boldsymbol{C}_{d_n}\bigr) \in \mathbb{R}^{n \times M}
\]

with $n$ being the total number of days in the time series $\boldsymbol{T}$. In practice, the daily measurements $\boldsymbol{C}_{d}$ is forecasted from the previous day $d-1$.
The primary objective of this work is to perform anomaly detection on such a daily production composing a given time series $\boldsymbol{T}$. An anomaly is defined as the overall intra-day sequence (i.e., window) of points of $\boldsymbol{T}$ itself, which deviates significantly from what is considered normal or expected behavior for EDF's power generation system for a day. This \textit{normality} is established based on patterns learned from historical data (including typical intra-day profiles for various day types). In practice, we consider 3 years of historical data.
Formally, in our industrial context, we define an anomaly detection function as follows: 

\begin{definition}[Anomaly Detection] We define an anomaly detection function $\mathcal{A}$ as $\mathcal{A} : \mathbb{R}^{n \times M} \times \mathbb{R}^{M} \longrightarrow [0,1]$,
that takes as input a time series $\boldsymbol{T} \in \mathbb{R}^{n \times M}$ and the last day $\boldsymbol{C}_{d_n} \in \mathbb{R}^M$, and returns a real-valued \textit{anomaly score} $S$ assessing how anomalous the last day is, given the historical context. A higher score indicates a higher likelihood of the last day being anomalous. 
\end{definition}

The function $\mathcal{A}$ can either correspond to a unique anomaly detection model, denoted by \(D\), or to an automatic detection solution, denoted by \(\mathcal{M}\), combining multiple detectors $D$. We will describe in the following sections the distinction between a unique detector and an automatic solution.
While our problem setting shares core principles with the broader literature on time series anomaly detection, such as the importance of temporal dependencies and seasonality, it presents several practical constraints limiting the applicability of a large panel of existing methods. These constraints are listed below.

\begin{itemize}
\item \textbf{(C1) Scalability}: The practical applicability of an anomaly detection system in an industrial context, such as managing energy grids, is heavily dependent on its computational performance. The system must exhibit low execution time to allow for rapid, proactive interventions.

\item \textbf{(C2) Limited hardware}: While model training can leverage high-performance computing resources, operational deployment must conform to strict hardware constraints. In our context, anomaly detection must run in near-real-time on standard CPU-based infrastructure, without access to GPUs or specialized accelerators.

\item \textbf{(C3) Interpretability}: In an operational context such as energy grid management, the value of an anomaly detection system is tightly coupled with its ability to produce interpretable outputs. Since detected anomalies must be reviewed and acted upon by domain experts (often under tight time constraints) models that function as black boxes are of limited utility.

\item \textbf{(C4) Data Diversity}: In our use case, time series are highly heterogeneous in terms of provenance and structure. First, missing values may have semantic meaning. Rather than removing them, one needs to preserve them to maintain data integrity and retrieve relevant knowledge that could indicate potential anomalies. Then, time series provenance (i.e., from diverse domains and measurement types) might lead to severe Out-of-Distribution scenarios, both in terms of trends and anomaly types.
\end{itemize}

\section{Time Series Anomaly Detection: \\\textit{Foundations and Background}}

In this section, we review existing time-series anomaly detection methods through the lens of the key operational constraints enumerated in the previous section.

\subsection{Raw-based time series anomaly detection}

In recent years, significant research has been conducted in the field of time-series anomaly detection. Numerous studies and experimental benchmarks have been written to summarize and analyze state-of-the-art methods \cite{Paparrizos22,Schmidl22,Boniol22,Wenig22,10.14778/3551793.3551830}. These comprehensive evaluations reveal a diverse landscape of methodologies (i.e., detectors $D$), which can be broadly classified into several foundational families based on their core operating principles~\cite{boniol2024divetimeseriesanomalydetection}. Figure~\ref{fig:taxonomy} depicts a large panel of methods grouped in process-centric families according to~\cite{boniol2024divetimeseriesanomalydetection}. 

One major family consists of \textbf{prediction-based methods}, which train a model on a given self-supervised prediction task. An observation is classified as anomalous if it significantly deviates from the model's prediction, with the anomaly score derived from the prediction error. Such methods can be \textit{reconstruction-based}~\cite{10.1145/2689746.2689747} or \textit{forecasting-based}~\cite{malhotra_long_2015}.

A second prominent category is \textbf{distance-based methods}, which operate under the assumption that anomalous points are isolated from the bulk of the data. These techniques identify outliers by measuring the distance of a data point to its nearest neighbors, with large distances indicating abnormality \cite{Knorr98}.

Closely related are \textbf{density-based methods}, which use a representation of the time series (i.e., tree~\cite{4781136} or graph structure~\cite{Series2GraphPaper}) and compute an anomaly score based on some density criteria of the given representation (i.e., depth of the tree, or centrality in the graph). 

While prediction-based methods are appealing due to their ability to model temporal dynamics, they often rely on complex architectures with high computational and memory demands, and may be ill-suited for specific use cases on high-velocity data with real-time deployment on constrained hardware requirements (i.e., failing to meet \textbf{C1} and \textbf{C2}). Additionally, their black-box nature poses challenges for interpretability (i.e., not respecting \textbf{C3}). Conversely, raw-based approaches such as distance- and density-based methods may offer lighter computation, but they struggle to handle data quality issues like missing values (failing to meet \textbf{C4}) and often lack the contextual transparency needed for non time series practitioners (i.e., not fully respecting \textbf{C3}). These limitations underscore the need for methods explicitly designed with industrial constraints in mind.

\subsection{\textbf{Lack of interpretability}: \textit{Toward Feature-based approaches}}

Feature-based methods offer a promising avenue for addressing the industrial constraints of interpretability, scalability, and robustness to data quality issues. By transforming raw time series into structured feature vectors (often composed of statistical, structural, or domain-specific descriptors), these approaches enable the use of appropriate density-based algorithms in a lower-dimensional, and meaningful space. Unlike raw-based distance- or density-based methods, feature-based techniques allow better handling of missing values through feature design and provide clear interpretability by exposing which characteristics contribute to the anomaly. As such, they strike a balance between operational feasibility and detection performance, making them particularly well-suited for real-world deployment in our industrial contexts.

In practice, the feature-based methodology~\cite{Gharghabi23,Bonifati2022,Christ18} entails a two-stage process that transforms the complex temporal problem into a more manageable static outlier-detection task. First, a raw time-series window is converted into a fixed-length vector of numerical features. These features can capture a wide range of properties, including basic statistics (mean, variance), frequency-domain information, and entropy measures. In the second stage, a classical outlier-detection algorithm is applied to the feature space to identify feature vectors that are abnormal relative to the rest of the population \cite{chalapathy2019deep}. Common algorithms used for this purpose include Isolation Forest~\cite{4781136}, Local Outlier Factor (LOF)~\cite{Breunig00}, and One-Class SVM~\cite{10.5555/3009657.3009740}.

\begin{figure}
    \centering
    \includegraphics[width=\linewidth]{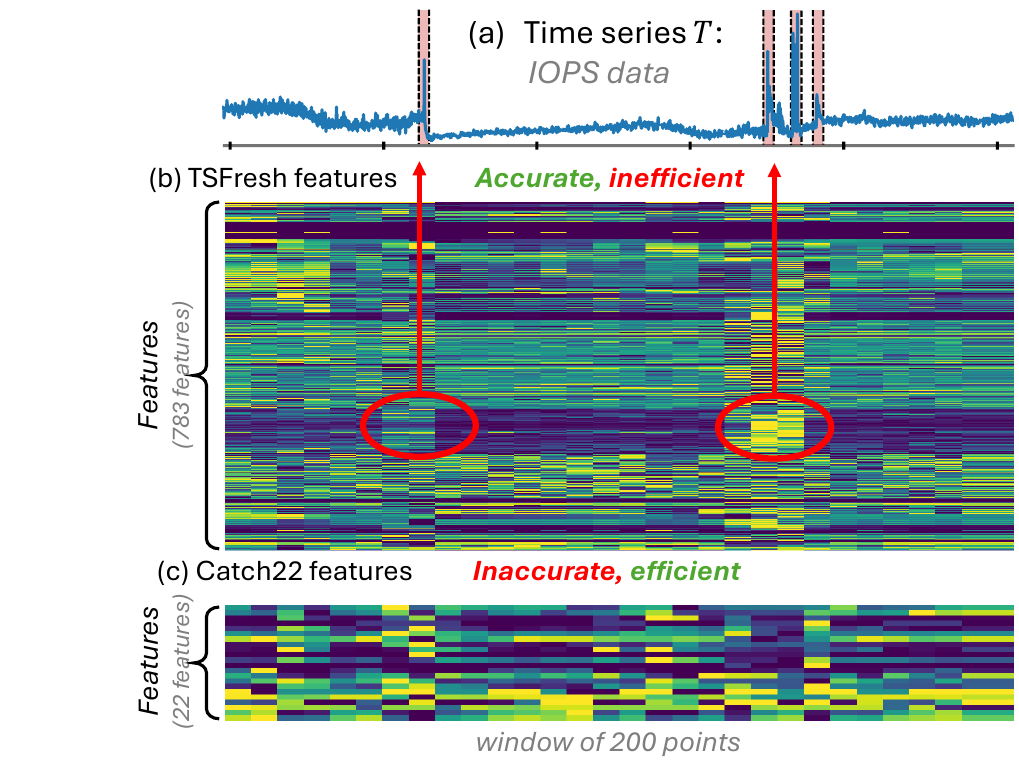}
    \caption{Example of \textsc{TSFresh} and \textsc{Catch22} features on IOPS~\cite{IOPS} time series.}
    \label{fig:feature-example}
\end{figure}

Several tools have been proposed for automatic feature extraction. Among the most widely used are HCTSA \cite{Fulcher17}, \textsc{TSFresh} \cite{Christ18}, and \textsc{Catch22} \cite{Lubba19}. HCTSA (Highly Comparative Time Series Analysis) performs massive feature extraction, computing over 7,700 features for each time series. These features span statistical distribution measures (e.g., mean, skewness, outlier ratios), autocorrelation and spectral properties, entropy metrics, and model-based characteristics.

\begin{figure*}
    \centering
    \includegraphics[width=\linewidth]{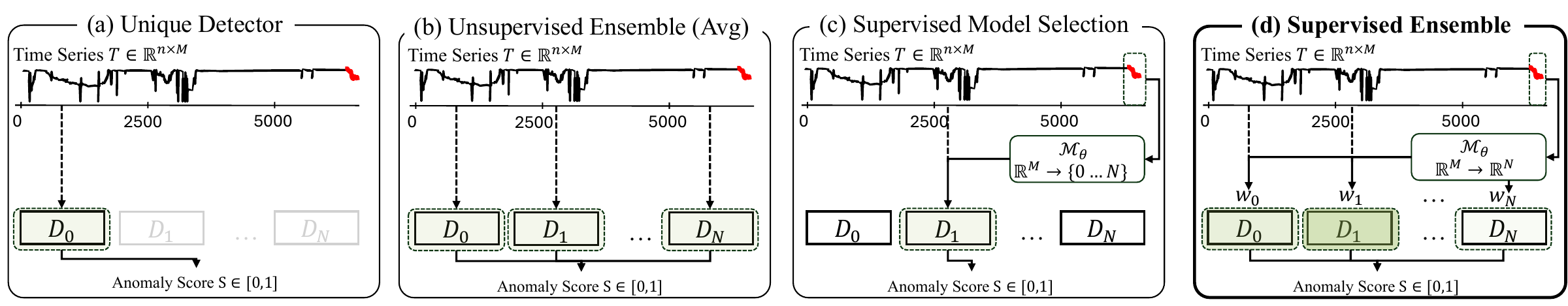}
    \caption{Time series anomaly detection: from unique detector (a) to supervised ensembling (d).}
    \vspace{-0.3cm}
    \label{fig:automatic}
\end{figure*}

\textsc{TSFresh} extracts 794 time series features by default and includes an automated feature selection process based on hypothesis testing. This framework supports both exploratory analysis and integration into operational pipelines, making it a flexible option for industrial applications.

\textsc{Catch22} builds upon HCTSA by identifying a compact subset of 22 features that provide strong classification performance across 93 UCR/UEA datasets~\cite{dau2019ucrtimeseriesarchive}. These features are selected to minimize redundancy and maximize diversity, but are specifically tuned for z-normalized time series (zero mean and unit variance) and classification tasks.

While each of these libraries provides valuable insights, they present specific limitations in our context: (i) HCTSA and \textsc{TSFresh} produce \textbf{high-dimensional feature sets, which are impractical} for direct use in real-time anomaly detection. Moreover, \textsc{TSFresh}'s selection is not tailored to anomaly-specific tasks, potentially misaligning with our deployment needs. (ii) \textsc{Catch22}, due to its compact size and design for rapid analysis, appears promising for initial exploration. However, its features were \textbf{optimized for classification, not anomaly detection}. 
Figure~\ref{fig:feature-example} depicts the \textsc{TSFresh} and \textsc{Catch22} features computed on an IOPS~\cite{IOPS} time series. TSfresh contains features that allow the detection of anomalies (labeled in red in Figure~\ref{fig:feature-example}), while \textsc{Catch22} features are not sufficient for detecting the same anomalies.

Given these constraints, no existing feature set is suitable for our industrial use case, underscoring the need for feature selection and evaluation tailored to our operational context.

\subsection{\textbf{No Universal Best}: \textit{Toward Automatic Solutions}}
\label{sec:automaticsolution}

Recent benchmarking studies have consistently shown that there is no single anomaly detection algorithm that performs best across highly heterogeneous collections of time series \cite{Paparrizos22,Schmidl22,10.14778/3611479.3611536, 10.5555/3737916.3741353}. Instead, the performance of individual methods varies significantly depending on the characteristics of the time series (e.g., stationarity, periodicity) and the nature of the anomalies (e.g., point-wise, contextual, or subsequence).

This observation is particularly important in our context, where we must monitor a diverse set of time series with varying behaviors and noise levels. Furthermore, while most research algorithms produce anomaly scores for each time point, our setting requires a single decision at the last timestamp, introducing an additional mismatch between standard benchmarks and operational needs.

To overcome these limitations, we use several automatic solutions (i.e., meta model and ensembling methods $\mathcal{M}$). Overall, two main strategies have been proposed: ensembling (supervised or unsupervised) and automatic model selection. Figure~\ref{fig:automatic} illustrates these strategies. Ensembling combines the outputs of multiple detectors, typically by averaging or maximizing their scores, to improve robustness. These methods often outperform individual algorithms on public benchmarks~\cite{10.14778/3611479.3611536}, but at the cost of higher computational overhead, which is problematic for real-time applications.

On the other hand, recent AutoML-based approaches have explored model selection as an alternative. These methods aim to select the most appropriate detector for each time series by learning a meta-model that maps extracted features to the best-performing method \cite{zhao2021automatingoutlierdetectionmetalearning, goswami2023unsupervisedmodelselectiontimeseries, 10.14778/3611479.3611536}. While promising, these techniques suffer from limited generalization in out-of-distribution settings~\cite{10.14778/3611479.3611536}. The latter is an essential concern for real-world deployment, where new time series may exhibit previously unseen patterns or behaviors.

Given these limitations, we need to adopt an ensembling strategy that prioritizes robustness in out-of-distribution settings while maintaining low computational cost. 

\section{Research Questions}
\label{sec:research_questions}

In light of the industrial constraints identified and the literature review in the previous section, our proposed solution is guided by several key research questions:

\begin{itemize}

\item \textbf{R1. How to combine detectors for optimal accuracy–efficiency trade-offs?} 
While ensembling can enhance robustness and detection accuracy, it also increases computational cost (C1). Conversely, model selection offers greater efficiency but may suffer in Out-of-Distribution scenarios (C4).

    \item \textbf{R2. Which feature set to use?} 
    A smaller feature set can improve scalability (C1, C2). But, small or non-dedicated features may negatively impact detection accuracy, reducing the system's practical usefulness (C3, C4).

    \item \textbf{R3. What is the impact of Out-of-Distribution?} 
    Identifying the correct automatic solution setting can keep execution time low (C1,C2) while providing greater robustness in Out-of-Distribution scenarios (C4).
\end{itemize}

These research questions directly inform the design of our \textsc{TAMIS} framework (detailed in the following section), ensuring that it balances operational feasibility with high detection performance. Overall, this study not only describes a data-based pipeline for anomaly detection in production but also presents a roadmap for moving from research findings in data mining and data management to production-based systems.

\section{Proposed Method: The \textsc{TAMIS} System}
\label{sec:ProposedApp}

\begin{figure*}
    \centering
    \includegraphics[width=\linewidth]{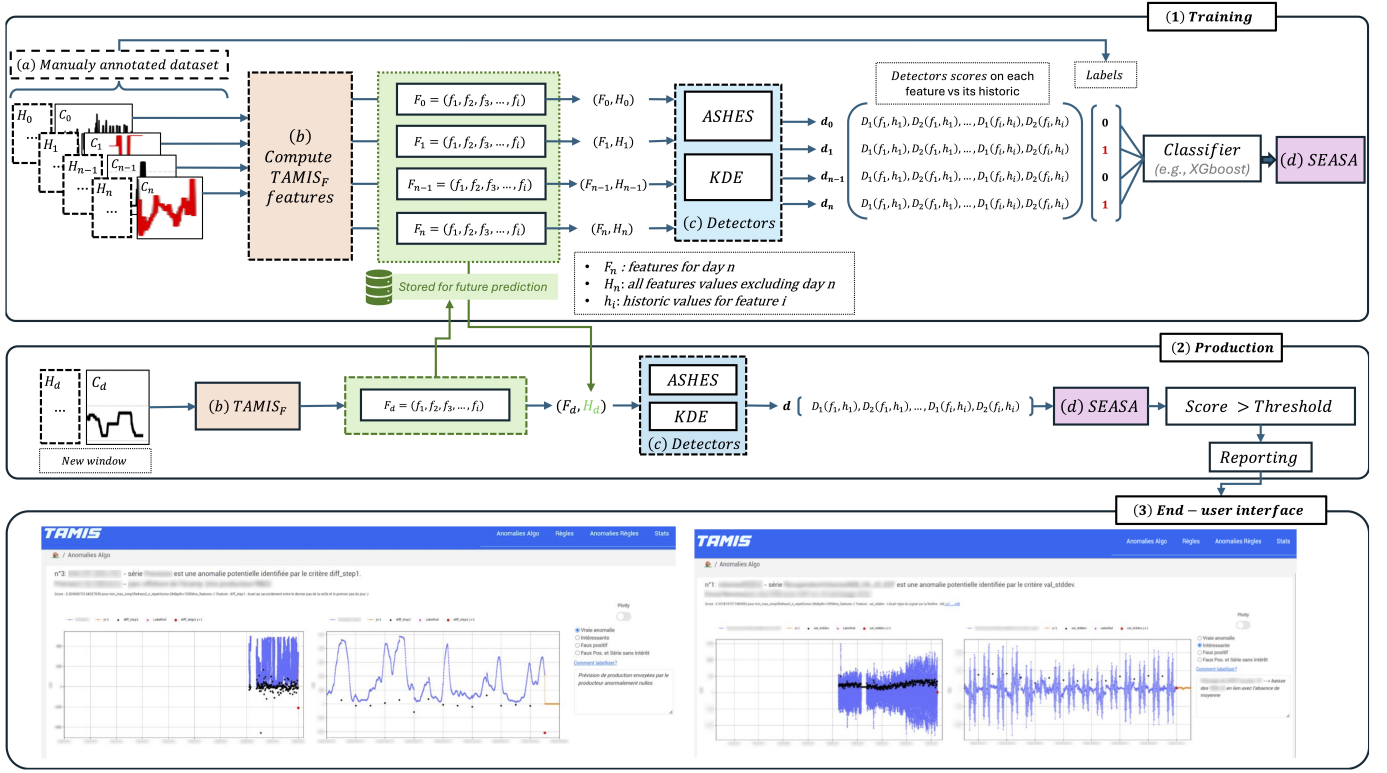}
    \caption{(1) Training and (2) production pipeline of the proposed solution \textsc{TAMIS}. The end-user interface is illustrated in (3).}
    \vspace{-0.3cm}
    \label{fig:pipeline}
\end{figure*}

In this section, we introduce \textsc{TAMIS}, a lightweight, interpretable anomaly detection system specifically designed for production-grade deployment. As shown in Figure~\ref{fig:pipeline}, the method involves a training phase composed of several steps: starting from manually labeled datasets (\textbf{step~(a)}), a novel feature set, \textsc{TAMIS}$_\mathcal{F}$, is computed for all available time series (\textbf{step~(b)}). Two complementary detectors are applied to these features (\textbf{step~(c)}), and their outputs are combined through a supervised ensemble, referred to as \textsc{SEASA} (\textbf{step~(d)}). Once trained, this supervised ensemble model is deployed for inference in production (Figure~\ref{fig:pipeline}(2)).

An important practical consideration is that precomputed features from the previous day are stored in a dedicated database and reused when computing anomaly scores for the following day. This caching mechanism drastically reduces runtime and enables near real-time performance in production.

Finally, the most abnormal time series are presented to the end-user through a visual interface, shown in Figure~\ref{fig:pipeline}(3). In the following sections, we detail the design of the proposed feature set \textsc{TAMIS}$_\mathcal{F}$, the two individual detectors, and the supervised ensembling mechanism, \textsc{SEASA}.

\subsection{TAMIS$_{\mathcal{F}}$: A Novel Feature Set}

The proposed feature set, \textsc{TAMIS}$_{\mathcal{F}}$, aims to capture the most informative and generalizable characteristics of industrial time series for anomaly detection. The feature extraction pipeline begins by aggregating raw measurements into daily windows, denoted as $\boldsymbol{C}_{d}$, from which a feature vector $\boldsymbol{F}_{d}$ is computed. The final feature set results from the integration of two complementary design strategies: (\textit{i}) an expert-based feature pool derived from domain knowledge, and (\textit{ii}) a TSAD-based feature pool identified through synthetic data generation and supervised feature selection.

\begin{table*}[tb]

\setlength{\tabcolsep}{7pt}

\begin{center}

\resizebox{\linewidth}{!}{%
\begin{tabular}{lcc}
\toprule
\textbf{Feature Description} & \textbf{Formula} & \textbf{Complexity} \\
\midrule
\rowcolor{gray!20}
\multicolumn{3}{c}{\textbf{Expert-based}: \textit{Top-10 features from energy-production expert-knowledge}} \\ 
\midrule
\textbf{amplitude:} maximum amplitude for a given window $\boldsymbol{C}_{d}$. & $\max(\boldsymbol{C}_{d}) - \min(\boldsymbol{C}_{d})$ & O(n)\\
\textbf{val max:} Maximum value in $\boldsymbol{C}_{d}$. & $\max(\boldsymbol{C}_{d})$ & O(n) \\
\textbf{val min:} Minimum value in $\boldsymbol{C}_{d}$. & $\min(\boldsymbol{C}_{d})$ & O(n)\\
\textbf{val mean:} Mean of $\boldsymbol{C}_{d}$. & $ \mu(\boldsymbol{C}_{d}) $ & O(n)\\
\textbf{val std:} Standard deviation of $\boldsymbol{C}_{d}$. & $\sigma(\boldsymbol{C}_{d})$ & O(n)\\
\textbf{missing values:} Count of missing (NaN) values in $\boldsymbol{C}_{d}$. & $\sum_{c_m \in\boldsymbol{C}_{d}} \mathbb{I}(c_m)$ & O(n)\\
\textbf{diff max:} Maximum consecutive difference in $\boldsymbol{C}_{d}$. & $ \max_{c_m \in\boldsymbol{C}_{d}}|c_m - c_{m-1}|$ & O(n)\\
\textbf{diff min:} Minimum consecutive difference in $\boldsymbol{C}_{d}$. & $ \min_{c_m \in\boldsymbol{C}_{d}}|c_m - c_{m-1}|$ & O(n)\\
\textbf{diff mean:} Average consecutive differences in $\boldsymbol{C}_{d}$. & $\frac{1}{|\boldsymbol{C}_{d}|-1} \sum_{c_m \in \boldsymbol{C}_{d}} |c_m - c_{m-1}|$ & O(n)\\
\textbf{diff step1:} difference between $c_0$ and $c^{\prime}_M$, first and last values of $\boldsymbol{C}_{d}$ and $\boldsymbol{C}_{d-1}$ respectively. & $ c_0 - c^{\prime}_M$ & O(n)\\
\midrule
\rowcolor{gray!20}
\multicolumn{3}{c}{\textbf{TSAD-based}: \textit{Top-10 features from a synthetic time series anomaly detection evaluation}} \\
\midrule
\textbf{sum reoccurring values:} Sum of all values that appear more than once in $\boldsymbol{C}_{d}$. & $\sum_{v \in V_{\text{reocc.}}}v$ & O(n) \\
\textbf{Benford correlation:} Correlation with the expected Benford's Law distribution of first digits. & $\text{corr}(P_{\text{obs}}, P_{\text{Benford}})$ & O(n) \\
\textbf{Fourier entropy:} Shannon entropy of the Power Spectral Density (PSD), with 100 bins. & $-\sum {p}_{i}\log_2({p}_{i})$ & O(n $\log(n)$) \\
\textbf{longest strike above mean:} Length of the longest consecutive run of values above the mean ($\mu$). & $\text{Length of run}>\mu$ & O(n) \\
\textbf{mean change:} Mean of the absolute differences between consecutive values. & $\text{mean}(|\Delta\boldsymbol{C}_{d}|)$ & O(n) \\
\textbf{last location of max:} Relative index of the last occurrence of the maximum value. & $\text{argmax}_{\text{last}}(\boldsymbol{C}_{d})/N$ & O(n) \\
\textbf{variation coefficient:} Ratio of the standard deviation ($\sigma$) to the mean ($\mu$). & $\sigma/\mu$ & O(n) \\
\textbf{permutation entropy:} Complexity measure based on ordinal patterns (dim=5, lag=1). & $-\sum_{i\in\{1,\ldots,D!\}}p_i\,\log_2(p_i)$  & O(n) \\
\textbf{has duplicate min:} Binary indicator for whether the minimum value appears more than once. & $\mathbb{I}(\text{count}(\min(\boldsymbol{C}_{d}))>1)$ & O(n) \\
\textbf{longest strike below mean:} Length of the longest consecutive run of values below the mean ($\mu$). & $\text{Length of run}<\mu$ & O(n) \\

\bottomrule
\end{tabular}
}

\captionsetup{justification=centering}
\caption{TAMIS$_{\mathcal{F}}$ features. Each feature is computed for an entire window (i.e., day $d$). Note: $\mathbb{I}(p)$ is the indicator function for NaN values. $\Delta {C}_{d}$ represents first-order differences. $V_{\text{reocc.}}$ is the set of reoccurring values in $C_{D}$.}
\vspace{-0.5cm}
\label{tab:features-TAMIS}

\end{center}
\end{table*}

\vspace{1em}
\subsubsection{\textbf{Expert-Based} Feature Pool}

The initial stage of \textsc{TAMIS}$_{\mathcal{F}}$ construction relied on domain expertise from energy production specialists. A total of 36 candidate indicators were manually proposed to describe operational behavior and potential deviations. To isolate the most relevant predictors, a structured feature selection protocol was implemented, combining ablation analysis and individual feature evaluation.

Specifically, three model configurations were trained to assess each feature’s contribution:
(i) A \textbf{baseline model} trained on the full feature set; (ii) An \textbf{ablation model} trained on all features except the one under consideration (leave-one-out); (iii) An \textbf{individual model} trained on the single feature alone.

Each configuration produced four recall-based performance curves, quantifying the ratio of correctly detected anomalies with respect to both the detection score and rank. By qualitatively and quantitatively analyzing these curves, the ten most informative and discriminative features were retained. These are reported in the \textit{expert-based} section of Table~\ref{tab:features-TAMIS}.

Nevertheless, two limitations arise with the approach mentioned above. First, since the selection relies on historically labeled anomalies, it may be biased toward previously observed anomaly types, potentially limiting its generalization to unseen fault patterns. Second, the anomaly distribution within the evaluation dataset may not reflect their true operational frequency, introducing sampling bias. As a result, the selected features might overemphasize frequently occurring anomalies at the expense of rare but critical ones.

\vspace{1em}
\subsubsection{\textbf{TSAD-Based} Feature Pool}

To address the scarcity of labeled data and improve cross-domain generalization, we developed a data-driven feature discovery pipeline inspired by time series anomaly detection (TSAD) research. This methodology converts the unsupervised detection problem into a supervised learning framework through synthetic data generation, followed by a two-stage feature selection process.

\noindent\textbf{Synthetic Data Generation and Feature Engineering:}
A labeled corpus is first synthesized by injecting artificial anomalies into normal time series (from \textsc{JO} dataset described in Section~\ref{sec:Datasets}). Inspired by recent experimental evaluation studies~\cite{Paparrizos22,Schmidl22}, these anomalies emulate representative fault types, including point perturbations, temporal shifts, and amplitude scaling. The resulting dataset provides explicit ground-truth labels for controlled feature selection. A high-dimensional feature space is then constructed by computing descriptive statistics and signal transformations over each daily window, integrating features from \textsc{TSFresh} and \textsc{Catch22}.

\noindent\textbf{Two-Stage Supervised Feature Selection:}
From the collected features, we perform the following selection pipeline:

\noindent\textit{Relevance Filtering:} Each feature is ranked according to its discriminative capacity for the synthetic labels, evaluated using p-value and Benjamini Hochberg procedure \cite{benjamini1995}, ANOVA F-test \cite{fisher1970statistical}, and gradient boosting importance scores \cite{Chen_2016}.

\noindent\textit{Redundancy Removal:} The top-ranked features are clustered based on pairwise correlation to remove redundant predictors. From each cluster, a single representative feature is retained by maximizing its target correlation (via ANOVA F-test score) while minimizing its average inter-feature correlation.

The ten features retained through this supervised selection process constitute the \textit{TSAD-based} component of \textsc{TAMIS}$_{\mathcal{F}}$, as detailed in Table~\ref{tab:features-TAMIS}. We evaluate the relevance of \textsc{TAMIS}$_{\mathcal{F}}$ in Section~\ref{sec:evalfeatures}. 
Note that the proposed feature set is not restricted to use within our anomaly detection framework, \textsc{TAMIS}. Our feature set can be integrated into any anomaly detection system, whether supervised or unsupervised. We evaluate the performance of traditional anomaly detection methods applied on \textsc{TAMIS}$_{\mathcal{F}}$ in Section~\ref{sec:evalOverall}.

\subsection{Anomaly Detectors Pool}
\label{sec:detectorpool}

This section describes the two anomaly detectors that compose the core of the \textsc{TAMIS} ensemble. 
The choice of using only two base detectors is deliberate, as it significantly reduces computational overhead while maintaining high detection accuracy and scalability in production environments.

Let $H_i$ denote the set of historical feature values for a given time series, and let $f_i \in \boldsymbol{F}_{d}$ represent the new observation under evaluation. 
Each detector outputs an anomaly score, which is subsequently combined by the supervised ensemble model described in Section~\ref{sec:SEASA}.

\vspace{1em}
\subsubsection{\textbf{\textsc{KDE}}: Kernel Density Estimation-based anomaly detection}

The first detector models the distribution of historical feature values using a univariate Gaussian kernel density estimator (\textsc{KDE})~\cite{10.1214/aoms/1177704472}. 
Anomalies are identified as points with low estimated density under this model.

Formally, let $\hat{d}_{H_i}$ denote the kernel density estimate fitted on $H_i$. 
To ensure numerical stability and produce a dimensionless score, the estimated density $\hat{d}_{H_i}(f_i)$ is rescaled by the empirical standard deviation of the historical data, $\hat{\sigma}(H_i)$. 
The resulting anomaly score is defined as follows:
\begin{equation}
\small
D_{\text{KDE}}(f_i, H_i) = 1 - \hat{d}_{H_i}(f_i)\,\hat{\sigma}(H_i)
\label{eq:kde}
\end{equation}

This normalization preserves the monotonic relationship between the anomaly score and the estimated density (i.e., higher scores indicate rarer events) while compensating for scale differences across heterogeneous time series. 
Because $D_{\text{KDE}}$ is a monotone decreasing function of $\hat{d}_{H_i}(f_i)$, ranking by $D_{\text{KDE}}$ is equivalent to ranking by $-\hat{d}_{H_i}$.

\vspace{1em}
\subsubsection{\textbf{\textsc{ASHES}}: Anomaly Scoring-based on Historical ExtremeS}

The second detector, denoted \textsc{ASHES} (\textit{Anomaly Scoring based on Historical ExtremeS}), quantifies how much a new observation $f_i$ extends beyond the historical data extremes. 
It combines two complementary components: an \emph{extremity rank} ($K$) and a \emph{relative amplitude ratio} ($r$), which are integrated into a single anomaly score.

\noindent\textbf{Extremity Rank ($K$):}
The extremity rank quantifies the position of $f_i$ within the ordered historical distribution.
It is defined as the minimum rank of $f_i$ when inserted into $H_i$ sorted in ascending and descending order. Formally:
\begin{equation}
\small
    K(f_i) = \min\big(\text{rank}_{\text{asc}}(f_i, H_i),\, \text{rank}_{\text{desc}}(f_i, H_i)\big)
    \label{eq:rank_k}
\end{equation}
A rank of $K = 1$ indicates that $f_i$ is a new global minimum or maximum. 
As a practical heuristic, when $K > 10$, the point is considered non-extreme and its anomaly score is set to zero.

\noindent\textbf{Relative Amplitude Ratio ($r$):}
To quantify the magnitude of deviation, we compute a relative amplitude ratio $r(f_i)$ that measures how much $f_i$ expands the range of the filtered historical set.
Let $H_{i_k}$ denote $H_i$ with the $K-1$ largest and $K-1$ smallest values removed, and let $\min_K = \min(H_{i_k})$, $\max_K = \max(H_{i_k})$. Formally, $r(f_i)$ is defined as follows:
\begin{equation}
\small
    r(f_i) = \frac{\max_K - \min_K}{\max(f_i, \max_K) - \min(f_i, \min_K)}
    \label{eq:ratio_r}
\end{equation}
A value of $r$ close to 1 indicates that $f_i$ lies within the typical range of the filtered extrema, while smaller values correspond to more substantial deviations from the historical amplitude.

\noindent\textbf{\textsc{ASHES} Anomaly Score:}
The anomaly score of \textsc{ASHES} integrates both $K$ and $r$ to capture the effect of extremity and deviation magnitude. 
Formally, it is computed as follows:
\begin{equation}
\small
    D_{\text{ASHES}}(f_i, H_i) = 0.5^{K-1} - r \times 0.5^{K}
    \label{eq:final_score}
\end{equation}
This exponentially weighted formulation ensures that points with low ranks (high extremity) and large deviations (small $r$) receive higher anomaly scores. 
Consequently, \textsc{ASHES} effectively highlights rare and impactful events that substantially alter the temporal dynamics of the monitored time series.

Overall, while the \textsc{KDE} detector provides a smooth, probabilistic view of data rarity, \textsc{ASHES} focuses on extreme deviations. 
Their complementary nature (density modeling versus rank-based extremity analysis) motivates their joint use within the supervised ensemble framework of \textsc{TAMIS}.

It is important to note that standard methods like HBOS~\cite{Goldstein2012HistogrambasedOS} or LOF~\cite{Breunig00} treat rarity and amplitude deviation equally. In our industrial context, a value can be \emph{rare} without being operationally critical if its relative amplitude is low. ASHES is specifically designed to address this by weighing the Extremity Rank ($K$) against the Relative Amplitude Ratio ($r$), offering a nuanced detection capability that off-the-shelf algorithms lack.

\subsection{Supervised Ensemble Anomaly Scores Aggregation} 
\label{sec:SEASA}

The final stage of the proposed approach introduces \textsc{SEASA} (\textit{Supervised Ensemble Anomaly Scores Aggregation}), a stacking-based ensemble model that integrates the outputs of multiple base anomaly detectors (i.e., \textsc{KDE} and \textsc{ASHES}) into a unified and more accurate prediction. This ensemble mechanism leverages supervised learning to automatically infer optimal weightings and interactions among detector outputs, thereby enhancing overall robustness.

\vspace{1em}
\subsubsection{Base Anomaly Score Generation}
For each daily time series window (i.e., day $d$), a feature vector $\boldsymbol{F}_d$ is computed using \textsc{TAMIS}$_{\mathcal{F}}$ feature set. 
This representation serves as input to the two independent detectors, \textsc{KDE} and \textsc{ASHES}, which each produce a scalar anomaly score reflecting the degree of abnormality for that day. 
As mentioned in the previous section, these scores capture complementary aspects of the data distribution.
The resulting base scores constitute the input features for the \textsc{SEASA} meta-learning stage.

\vspace{1em}
\subsubsection{Meta-Learner and Prediction Strategy}
The second level of \textsc{SEASA} is a \textit{meta-learner} $\mathcal{M}$ designed to combine the raw anomaly scores from the base detectors into a final anomaly prediction.
In practice, we employ an XGBoost classifier~\cite{Chen_2016} as $\mathcal{M}$. 
Formally, given base detector outputs $\{D_{\text{KDE}}(f_i), D_{\text{ASHES}}(f_i)\}$ and corresponding labels $y_i \in \{0,1\}$, the meta-learner learns a mapping as follows:
\begin{equation}
\small
\hat{y}_i = \mathcal{M}\big(D_{\text{KDE}}(f_i), D_{\text{ASHES}}(f_i)\big)
\end{equation}
where $\hat{y}_i$ denotes the final anomaly probability. 
This supervised aggregation strategy allows the ensemble to adaptively weight detector contributions according to their reliability across different anomaly types.

\vspace{1em}
\subsubsection{Experimental setup}
To obtain an unbiased prediction for every point in our dataset, we utilize a Leave-One-Out Cross-Validation (LOOCV) procedure. For each point being evaluated, the XGBoost model is trained on all other labeled days and then makes a prediction on the single point that was left out. This operation is repeated for the entire dataset, ensuring that every prediction is made on data not seen during the training of that specific model instance.
In deployment, the \textsc{SEASA} model is trained once using the available labeled dataset and then applied to new daily measurements.

\section{Experiments}
\label{sec:expEval}

This section presents a comprehensive evaluation of our proposed approach, \textsc{TAMIS}, which addresses the research questions outlined in Section~\ref{sec:research_questions}, and is organized as follows:

\begin{itemize}
    \item \textbf{Overall evaluation:} We begin by assessing the global performance of \textsc{TAMIS} (accuracy and throughput) compared to a comprehensive set of baselines. Within this overall evaluation, we analyze different strategies for combining anomaly detectors, focusing on whether it is more advantageous to learn a selection model or to use an ensemble approach. This experiment addresses \textbf{R1}.

    \item \textbf{Impact of the feature set:} We then investigate how different feature representations affect performance (i.e., accuracy and efficiency). We compare \textsc{TAMIS}$_{\mathcal{F}}$, against two widely used alternatives, \textsc{TSFresh}~\cite{Christ18} and \textsc{Catch22}~\cite{Lubba19}. This experiment addresses \textbf{R2}.
    
    \item \textbf{Out-of-distribution and transfer evaluation:} Finally, we evaluate the ability of \textsc{TAMIS} to generalize across different types of energy production systems. This experiment addresses \textbf{R3}.
\end{itemize}

\subsection{Experimental Setup}

All experiments are conducted on a single node with two Intel Xeon Gold 6234 CPUs at 3.30 GHz. Each node has 16 physical cores and 384 GB of RAM.
For reproducibility, we make our implementation publicly available~\footnote{Code Available at: \url{https://github.com/VautierNicolas/TAMIS}}. 

\subsubsection{Datasets}
\label{sec:Datasets}

In our industrial context, time series are highly diverse in nature and come from two areas: thermal and hydraulic, from EDF's thermal and hydraulic production facilities, respectively. This diversity leads to a wide variety of behaviors.
Moreover, several manual annotation campaigns have been conducted to obtain high-quality labels, resulting in a total of 6,931 annotated days with 376 anomalies.
We divide our benchmark in the three datasets described in Table~\ref{tab:datasets}. Overall, these datasets are as follows:

\begin{itemize}
    \item \textbf{\textsc{THERM} dataset}: contains time series and labeled windows from the thermal perimeter. More precisely, time series corresponds to (i) marginal costs, penalties, start-up, and operating costs (\euro{}$\mathrm{MW}^{-1}$ or \euro{}); (ii) demand, reference, and optimized programs ($\mathrm{MW}$); (iii) operating points for thermal units ($\mathrm{MW}$, gradient, number of modulations); and (iv) gas volumes (\(\mathrm{m^3}\)). In total, this set contains 4515 individual sensors.
    \item \textbf{\textsc{HYDRAU} dataset}: contains time series and labeled windows from the hydraulic perimeter. More precisely, time series corresponds to (i) minimum and maximum volumes and trajectories of reservoirs (\(\mathrm{hm^3}\)); (ii) minimum, maximum, and turbine flow rates of power plants (\(\mathrm{m^3\,s^{-1}}\)); and (iii) minimum, maximum, and scheduled power output of power plants ($\mathrm{MW}$). In total, this set contains 1740 individual sensors.
    \item \textbf{\textsc{JO} dataset}: contains both time series and labeled windows from thermal and hydraulic perimeters; the intersection of the \textsc{JO}, \textsc{THERM}, and HYDRAU is empty. The aim of \textsc{JO} is to represent the initial training database for monitoring systems at EDF. In total, this set contains 3198 individual sensors.
\end{itemize}

\begin{table}[tb]

\setlength{\tabcolsep}{4pt}

\begin{center}

\resizebox{\columnwidth}{!}{%
\begin{tabular}{lccc}
\toprule
Characteristics & \textsc{JO} & \textsc{HYDRAU} & \textsc{THERM} \\

\midrule
\textbf{Number of sensors}      & 3198      & 1740       & 4515 \\ 
\textbf{Total number of measurements}             & 219M & 267M  & 103M \\ 
\textbf{Number of Labeled days}    & 5813      & 540        & 578 \\ 
\texttt{NaN} \textbf{ratio}        & 20.3\%      & 3.5\%        & 27.9\% \\ 
Total number of \texttt{NaN}        &   55334     &   640       &  7827  \\ 
\textbf{Number of Anomalies}        & 307       & 57         & 12 \\ 
\bottomrule
\end{tabular}
}
\captionsetup{justification=centering}
\caption{Datasets characteristics. \texttt{NaN} ratio: number of labeled days with at least one \texttt{NaN} value.}
\vspace{-0.5cm}
\label{tab:datasets}

\end{center}
\end{table}

{
While existing benchmarks~\cite{Paparrizos22,10.5555/3737916.3741353,Wenig22} have driven significant progress, our dataset introduces distinct industrial challenges that are under-represented in the literature:
\begin{itemize}
    \item \textbf{Data Quality and Semantic NaNs:} Unlike curated benchmarks where missing values are rare, cleaned, or artificially imputed, our dataset retains the original data quality issues inherent to production environments. As shown in Table \ref{tab:datasets}, missing values occur at a maximum of 27.9\% of the labeled days. 
    \item \textbf{Forecast Validation Task:} Most public benchmarks focus on monitoring continuous raw sensor streams for retrospective anomaly detection. In contrast, our dataset addresses the validation of \textit{short-term production forecasts}. This task involves comparing a new value against historical profiles to anticipate future failures, a structure distinct from standard stream monitoring.
    \item \textbf{Physical Heterogeneity:} Spanning both thermal and hydraulic domains, our dataset covers a wider range of physical behaviors (e.g., flow rates, temperatures) and economic variables (e.g., marginal costs).
\end{itemize}
}

Finally, we release an anonymized version of our datasets~\footnote{Datasets: \url{https://github.com/VautierNicolas/TAMIS_dataset}}. To the best of our knowledge, this collection is the most extensive dataset of energy production time series for anomaly detection, making it a significant contribution of our work. This release aims to encourage future research in this direction.

\subsubsection{Baselines}

We compare our proposed approach, \textsc{TAMIS}, against several categories of existing methods. We begin by evaluating \textsc{TAMIS} against \emph{raw-based} detectors, i.e., approaches that operate directly on subsequences of the time series of interest. Next, we compare \textsc{TAMIS} to \emph{feature-based} detectors that use the same feature set as our solution. Finally, we benchmark \textsc{TAMIS} against a set of \emph{automatic solutions}. Overall, the baselines considered are the following:

\begin{itemize}
    \item \textbf{Raw-based Detectors:} We consider three representative methods: One-Class SVM~\cite{10.5555/3009657.3009740} (OCSVM), Histogram-Based Outlier Score~\cite{Goldstein2012HistogrambasedOS} (HBOS), and Local Outlier Factor~\cite{Breunig00} (LOF). These baselines are chosen based on the industrial constraints (cf. Sec~\ref{sec:Probdef}), and for their ability to operate on raw time series or extracted features.
    \item \textbf{Feature-based Detectors:} We employ the same three methods (OCSVM, HBOS, and LOF) applied to features. Additionally, we compare with \textsc{KDE} and our proposed feature-based method, \textsc{ASHES} (cf. Sec~\ref{sec:detectorpool}). As each detector outputs a vector of scores (i.e., one for each feature), the best aggregation method, among min, max, mean, and the meta-learner $\mathcal{M}$ (cf. Sec~\ref{sec:SEASA}), is applied.
    \item \textbf{Automatic solutions:} As individual detectors struggle to remain robust on heterogeneous datasets (cf. Sec~\ref{sec:automaticsolution}). We compare \textsc{TAMIS} with two automatic methods, including an unsupervised average ensemble (\textit{Avg Ens}) and the model selection strategy (\textit{MS}) introduced in~\cite{10.14778/3611479.3611536}.
\end{itemize}

\begin{figure*}[tb]
    \centering
    \includegraphics[width=\linewidth,scale=3]{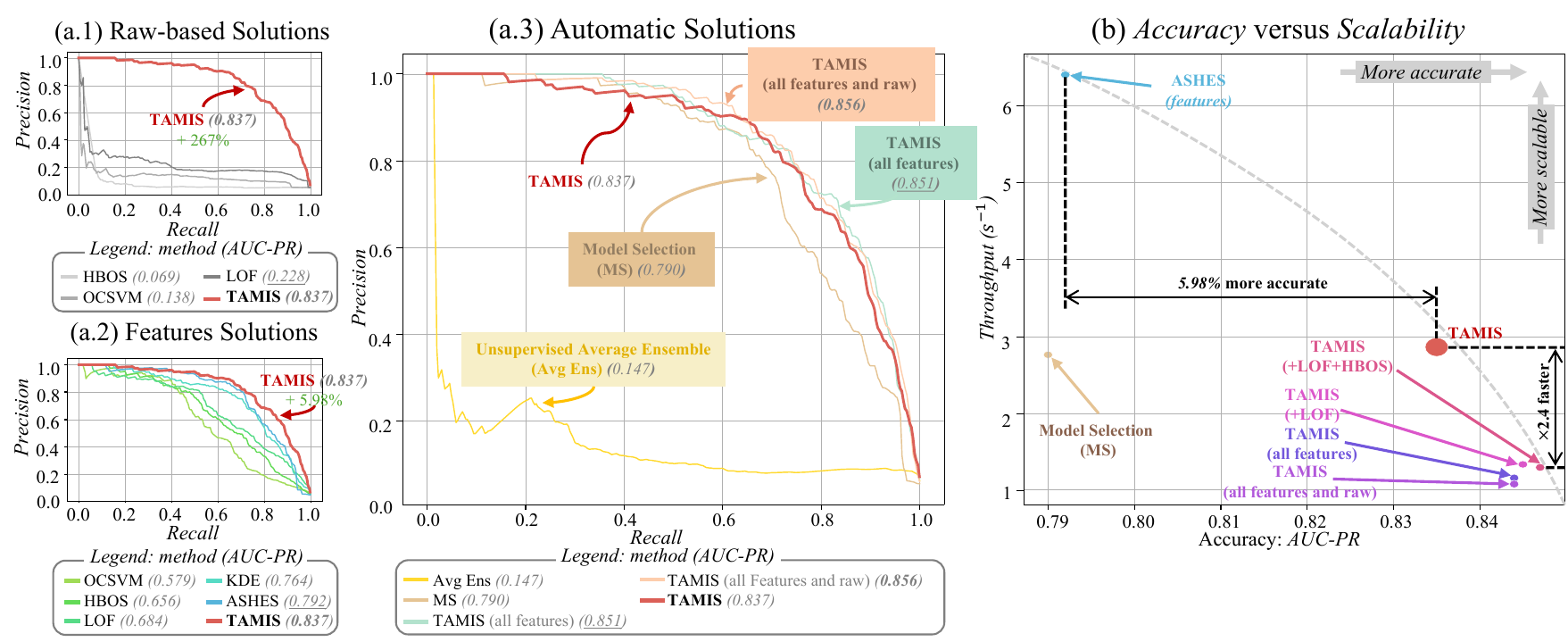}
    \caption{Precision-Recall curves of \textsc{TAMIS} against (a) detectors on raw data; (b) detectors on \textsc{TAMIS}$_{\mathcal{F}}$ feature set; (c) Automatic solutions (Model Selection and Ensembling). (d) Throughput versus AUC-PR for TAMIS against most competitive baselines.} 
    \vspace{-0.3cm}
    \label{fig:benchmark}
\end{figure*}

None of the baselines natively handles \texttt{NaN}s. Thus, we impute missing values using both forward and backward fill methods. This strategy is motivated by three considerations: (i) forward/backward filling maintains local temporal consistency, leading to more reliable anomaly scores; (ii) propagating the last (or next) observed value retains the piecewise-constant nature of the signal; and (iii) the imputed values remain within the true range of the observed data.

Finally, we evaluate several variants of our proposed approach. First, we consider a version of \textsc{TAMIS} that incorporates a larger set of feature-based detectors (referred to as \textsc{TAMIS (all features)}), combining OCSVM, HBOS, LOF, \textsc{KDE}, and \textsc{ASHES}. We also examine a more comprehensive configuration combining both feature-based and raw-based detectors (denoted \textsc{TAMIS (all features and raw)}), which, on top of the previously mentioned detectors, includes HBOS, OCSVM, and LOF on raw subsequences.

\subsubsection{Evaluation Measures}

Anomalies being rare, the Area Under the ROC curve (AUC-ROC) tends to overestimate the accuracy of detectors~\cite{10.5555/3737916.3741353}. We thus consider the area under the Precision-Recall curve (AUC-PR). 
Beyond accuracy, industrial constraints require low execution time. Thus, we use throughput to measure efficiency and scalability.
Note that more recent and robust evaluation measures exist, such as Volume Under the Surface (VUS)~\cite{10.14778/3551793.3551830}. Nevertheless, our problem is to determine whether a given day $D$ is abnormal (the score produced by our approach is a single value $S \in \mathbb{R}$). Therefore, our problem is not affected by potential misalignment, justifying the use of VUS.

\begin{figure*}
\centering
\centering
    \includegraphics[width=\linewidth]{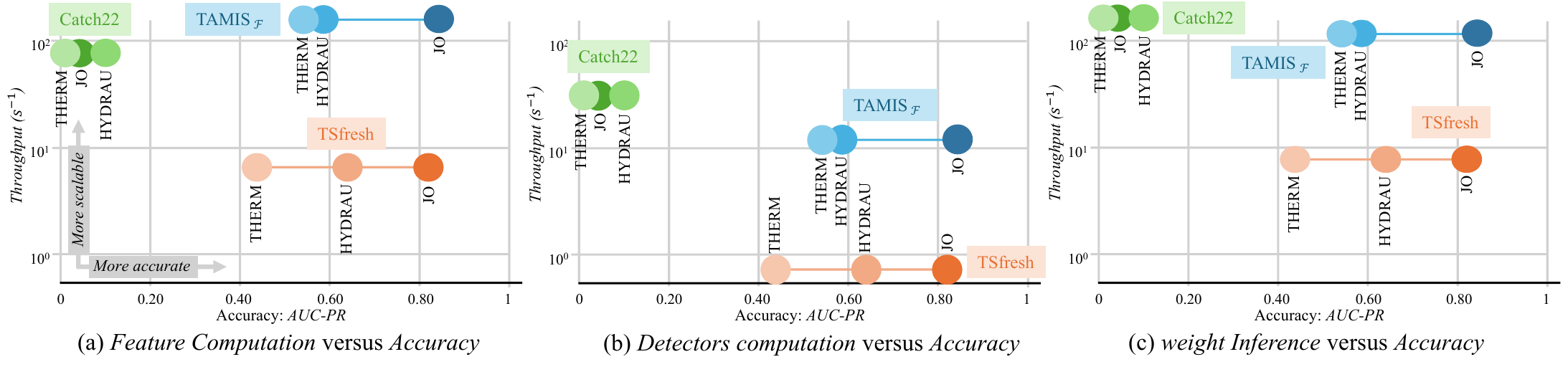}
\caption{Comparision of AUC-PR versus throughput between \textsc{TAMIS} using \textsc{TAMIS}$_\mathcal{F}$, \textsc{Catch22} and \textsc{TSFresh} as feature sets. The throughput corresponds to (a) Features computation, (b) Detectors computation, and (c) weights inference.}
\vspace{-0.3cm}
\label{fig:feature-eval}
\end{figure*}

\subsection{Overall Evaluation}
\label{sec:evalOverall}

\begin{table}[tb]

\setlength{\tabcolsep}{4pt}

\begin{center}

\resizebox{\columnwidth}{!}{%
\begin{tabular}{lccc}
\toprule
\textbf{Methods} & \textbf{JO} & \textbf{HYDRAU} & \textbf{THERM} \\

\rowcolor{gray!20}
\multicolumn{4}{c}{\textit{Detectors on raw time series}} \\ 

HBOS & 0.069 & 0.297 & 0.041 \\
LOF & 0.228 & 0.226 & 0.072 \\
OCSVM & 0.138 & 0.143 & 0.028 \\

\rowcolor{gray!20}
\multicolumn{4}{c}{\textit{Detectors on (TAMIS$_\mathcal{F}$ features)}} \\ 
HBOS & 0.656 & 0.596 & 0.191 \\
OCSVM & 0.579 & 0.527 & 0.156 \\
LOF & 0.684 & 0.523 & 0.452 \\
KDE & 0.764 & 0.554 & 0.462 \\
ASHES & 0.792 & 0.571 & 0.229 \\

\rowcolor{gray!20}
\multicolumn{4}{c}{\textit{Automatic Solutions (on TAMIS$_\mathcal{F}$ features)}} \\ 
Average Ensemble & 0.147 & 0.209 & 0.033 \\
Model Selection & 0.790 & 0.569 & 0.157 \\
\textsc{TAMIS (all features)} & \underline{0.851} & \underline{0.683} & 0.486 \\
\textsc{TAMIS (all features and raw)} & \textbf{0.856} & \textbf{0.691} & \underline{0.522} \\
\textsc{TAMIS} & 0.837 & 0.585 & \textbf{0.555} \\
\bottomrule
\end{tabular}
}

\captionsetup{justification=centering}
\caption{Accuracy (AUC-PR) of TAMIS and baselines applied on raw data and TAMIS$_\mathcal{F}$ features.}
\vspace{-0.5cm}
\label{tab:res_details}

\end{center}
\end{table}

In this section, we evaluate the global performance of \textsc{TAMIS} relative to all baseline configurations. The results are summarized in Table~\ref{tab:res_details} and Figure~\ref{fig:benchmark}.

\noindent\textbf{Raw time series as input:}
The first block of Table~\ref{tab:res_details} demonstrates that directly applying detectors to raw time series leads to poor accuracy. Across all datasets, AUC-PR values of LOF (i.e., the most accurate raw-based detector) remain below 0.23 on \textsc{JO} and \textsc{HYDRAU}, and drop to approximately 0.07 on \textsc{THERM}. This confirms that unprocessed time series lack the discriminative structure required for effective anomaly detection in electrical production systems.

\noindent\textbf{Feature-based detection:}
Performances increase when detectors operate on \textsc{TAMIS}$_{\mathcal{F}}$. Among individual detectors, \textsc{ASHES} achieves the highest AUC-PR on \textsc{JO} (0.792), while maintaining strong results on \textsc{HYDRAU} (0.571) and \textsc{THERM} (0.229). This confirms that the \textsc{TAMIS}$_{\mathcal{F}}$ feature set provides a more informative representation than raw subsequences.

\noindent\textbf{Automatic aggregation approaches:}
As shown in Table~\ref{tab:res_details}, supervised ensemble methods (i.e., \textsc{SEASA}) significantly outperform both unsupervised ensembles and supervised model selection. The largest ensemble (i.e., \textsc{TAMIS (all features and raw)}) achieves the best overall accuracy (AUC-PR = 0.856), surpassing the best individual detector (\textsc{ASHES}) by +6.40 points. These results highlight the strong complementarity between detectors and the benefits of \textsc{SEASA}.

\noindent\textbf{Accuracy–throughput trade-off:}
However, Figure~\ref{fig:benchmark} reveals that this accuracy improvement comes at the cost of computational efficiency. The largest ensemble has the lowest throughput because of its high inference cost. In contrast, the proposed \textsc{TAMIS} (i.e., using only two detectors, namely \textsc{KDE} and \textsc{ASHES}) achieves an optimal balance between accuracy and scalability. Specifically, it delivers near-best accuracy while being 2.4× faster, establishing \textsc{TAMIS} as the best trade-off solution for large-scale deployment.

\begin{tcolorbox}[colback=gray!5!white,colframe=gray!80!white,boxsep=2pt,top=2pt,bottom=2pt]
\textit{\textbf{Answer to R1:} Combining heterogeneous detectors improves anomaly detection accuracy, especially when integrating both raw-data-based and feature-based approaches. Supervised ensembles outperform unsupervised and selection-based strategies, while the proposed \textsc{TAMIS} configuration achieves the best accuracy–efficiency trade-off with only two detectors.}
\end{tcolorbox}

\begin{figure*}
    \centering
    \includegraphics[width=\linewidth]{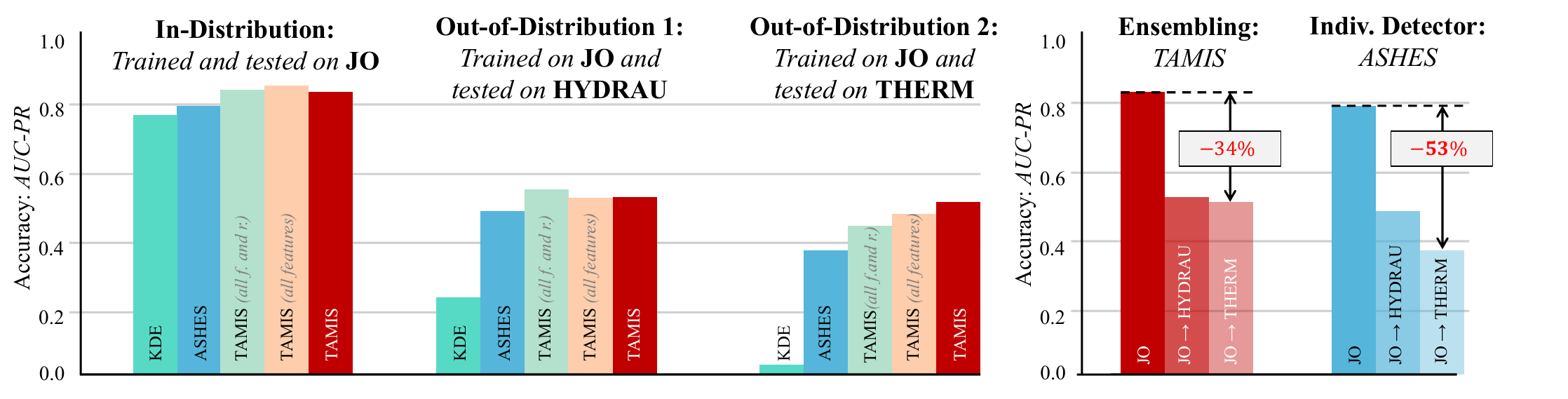}
    \caption{Comparison of our proposed approach \textsc{TAMIS} versus baselines on different models in-distribution (trained and tested on \textsc{JO} dataset) and two out-of-distribution scenarios. Note that \textit{all f. and r.} stands for \textit{all features and raw}.}
    \label{fig:ood_vs_id}
    \vspace{-0.3cm}
\end{figure*}

\subsection{Evaluating the Relevance of the Feature Sets}
\label{sec:evalfeatures}

Figure~\ref{fig:feature-eval} presents the trade-off between accuracy (AUC-PR) and throughput across the three stages of the \textsc{TAMIS} pipeline: (a) feature computation, (b) detector execution, and (c) weight inference. Three feature sets are compared: \textsc{TAMIS}$_\mathcal{F}$, \textsc{Catch22}, and \textsc{TsFresh}. 

\noindent\textbf{Features computation:}
Across all datasets, \textsc{TAMIS}$_\mathcal{F}$ provides the best trade-off between accuracy and efficiency. It reaches the highest AUC-PR values ($\approx$ 0.8–0.9) while sustaining throughputs several times higher than \textsc{TsFresh} and comparable to \textsc{Catch22}. While \textsc{Catch22} remains computationally lightweight, its AUC-PR saturates around 0.15, revealing limited expressiveness for anomaly detection in electrical production systems. Conversely, \textsc{TsFresh} reaches high accuracy, but at a prohibitively high computational cost.

\noindent\textbf{Pipeline-level efficiency:}
Figure~\ref{fig:feature-eval}(b) and (c) further demonstrate that the efficiency advantage of \textsc{TAMIS}$_\mathcal{F}$ persists through both the detector computation and ensemble weight inference stages. When all processing stages are considered, \textsc{TAMIS}$_\mathcal{F}$ maintains high accuracy and throughput. 

\begin{tcolorbox}[colback=gray!5!white,colframe=gray!80!white,boxsep=2pt,top=2pt,bottom=2pt]
\textit{\textbf{Answer to R2:} \textsc{TAMIS}$_\mathcal{F}$ feature set achieves the best accuracy–throughput balance among existing alternatives. It offers near-state-of-the-art detection accuracy at a significantly lower computational cost.}
\end{tcolorbox}

\subsection{Toward Production: an Out-of-Distribution test}

To assess the robustness and generalization capability of the proposed approach, we compare its performance under in-distribution (ID) and out-of-distribution (OOD) conditions. A baseline model is trained and tested on \textsc{JO}, representing the ID setting, while OOD performance was evaluated by applying the same model to the \textsc{HYDRAU} and \textsc{THERM} datasets, which differ in operational regimes and signal characteristics. 
Figure~\ref{fig:ood_vs_id} reports the AUC-PR results for three evaluation configurations: (i) in-distribution (trained and tested on \textsc{JO}), (ii) OOD~1 (\textsc{JO}~$\rightarrow$~\textsc{HYDRAU}), and (iii) OOD~2 (\textsc{JO}~$\rightarrow$~\textsc{THERM}). We include the best-performing individual detectors (\textsc{KDE} and \textsc{ASHES}) as baselines.

\noindent\textbf{In-distribution performance.}
All methods perform strongly (AUC-PR $\approx$ 0.8–0.9) in ID settings, demonstrating that both individual detectors and \textsc{TAMIS} successfully capture the statistical structure of the in-distribution data. These results establish a reference for subsequent OOD comparisons.

\noindent\textbf{Cross-domain transfer: \textsc{JO} $\rightarrow$ \textsc{HYDRAU}.}
In the first OOD scenario, performance decreases across all methods. \textsc{KDE} shows substantial degradation (AUC-PR $\approx$ 0.2), indicating sensitivity to distributional shifts. In contrast, \textsc{TAMIS} maintains a significantly higher AUC-PR ($>0.5$), underscoring its superior ability to generalize across domains with differing noise levels and dynamic properties. Note that in such a scenario, \textsc{ASHES} also shows strong performances, but still suffers from a larger drop than \textsc{TAMIS} between ID and OOD.

\noindent\textbf{Cross-domain transfer: \textsc{JO} $\rightarrow$ \textsc{THERM}.}
The second transfer scenario, from \textsc{JO} to \textsc{THERM}, is more challenging. All models experience additional performance loss, but \textsc{TAMIS} again remains the most robust, consistently outperforming \textsc{KDE} and \textsc{ASHES} by a large margin. This result highlights that the ensemble and feature-aggregation mechanisms within \textsc{TAMIS} effectively mitigate domain-specific overfitting.

\noindent\textbf{Relative degradation analysis.}
The right-hand plots in Figure~\ref{fig:ood_vs_id} quantify the relative AUC-PR loss between ID and OOD conditions. \textsc{TAMIS} exhibits only a $\sim$34\% decrease when tested on \textsc{THERM}, whereas \textsc{ASHES} experiences a 53\% drop. These findings confirm that the ensemble-based formulation of \textsc{TAMIS} acts as a regularizer, maintaining higher predictive stability under data distribution shifts.

\begin{tcolorbox}[colback=gray!5!white,colframe=gray!80!white,boxsep=2pt,top=2pt,bottom=2pt]
\textit{\textbf{Answer to R3:} \textsc{TAMIS} exhibits strong robustness under out-of-distribution conditions, retaining higher accuracy than individual detectors while maintaining low computational cost. The latter confirms the applicability of \textsc{TAMIS} in production, where distribution drift and heterogeneity across time series are expected.}
\end{tcolorbox}

{

\section{Operational Insights and Extensions}

This section discusses the operational feedback obtained from applying our approach in practice and outlines extensions toward more general benchmarks.

\subsection{Operational feedback}
The deployment of our proposed approach, TAMIS, provided valuable insights into the \emph{human-in-the-loop} requirements for industrial anomaly detection:

\noindent\textbf{Mitigating Alert Fatigue:} Before the introduction of our proposed approach, the main challenge was the volume of alarms. By introducing a ranking mechanism (based on a threshold strategy), TAMIS significantly reduced alert fatigue. Experts reported that the daily report enables more efficient triage by focusing only on the most critical deviations.

\noindent\textbf{The Value of Interpretability:} While deep learning approaches often act as black boxes, the choice of explicit features in $TAMIS_{\mathcal{F}}$ was critical for adoption. Operational teams validated that these features serve as immediate proxies for physical root causes, enabling faster decision-making.

\noindent\textbf{False Positives and Distribution Shifts:} The majority of false positives are triggered by Out-of-Distribution (OOD) events. However, our experiments in Figure~\ref{fig:ood_vs_id} demonstrate that TAMIS effectively reduces these errors compared to individual detectors, showing a relative resilience that is crucial for maintaining trust over time.
    

\noindent\textbf{From Generic to Tailored Monitoring:} A common industrial hurdle is the \emph{cold start} problem, where historical labels are unavailable for a new production unit. Our deployment strategy leverages TAMIS's transferability to address this. The system is initially deployed with a generic model pre-trained on available corporate data. The visualization interface then serves a dual purpose: (i) it supports daily decision-making and (ii) acts as a data annotation tool. As experts validate or reject alerts in the daily reports, they progressively build a high-quality, domain-specific labeled dataset. This feedback loop allows TAMIS to be iteratively retrained, refining the system's sensitivity to domain-specific characteristics.

\subsection{Generalization on public Benchmarks}
While TAMIS is designed to address specific industrial constraints (e.g., missing values, heterogeneity), it is crucial to verify that its performance is not solely due to the specificity of our use case and datasets. To assess the generalizability of our approach, we evaluated TAMIS on TSB-UAD~\cite{Paparrizos22}, a comprehensive, heterogeneous benchmark across domains.
We compared TAMIS against the strongest baselines identified in a recent evaluation of model selection for time series anomaly detection~\cite{10.14778/3611479.3611536}. More specifically, we consider the \textit{Best Detector} (i.e., NormA), the \textit{Unsupervised Average Ensemble}, and the \textit{Best Model Selection} (Best MS) strategy identified in~\cite{10.14778/3611479.3611536}.

Figure \ref{fig:tsb_uad} reports the Volume Under the Surface (VUS-PR) accuracy~\cite{10.14778/3551793.3551830} of TAMIS versus the baselines on TSB-UAD. We observe that TAMIS (VUS-PR $\approx$ 0.38) significantly outperforms the \textit{Average Ensemble} and the \textit{Best Detector}. Moreover, TAMIS outperforms the \textit{Best Model Selection} strategy (VUS-PR$\approx$ 0.35).
This result confirms that TAMIS captures fundamental anomaly characteristics that generalize well to diverse domains beyond energy production.

\begin{figure}
    \centering
    \includegraphics[width=\linewidth]{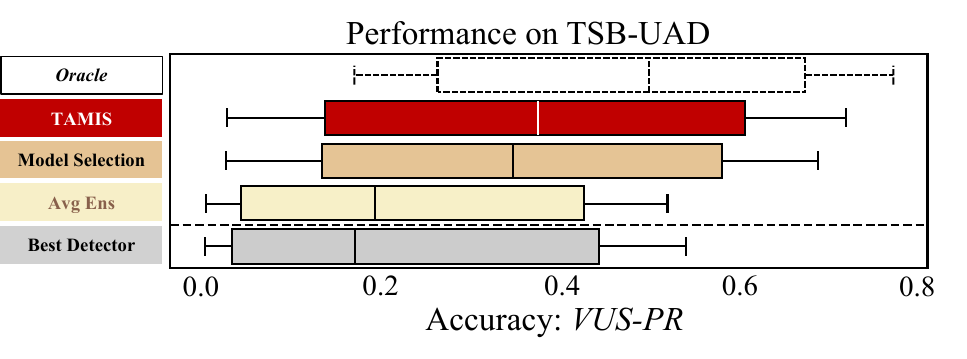}
    \caption{TAMIS vs Best Detector, Avg Ens, the Best Model Selection strategy and the theoretical Oracle on TSB-UAD~\cite{Paparrizos22}.}
    \label{fig:tsb_uad}
    \vspace{-0.3cm}
\end{figure}

}

\section{Conclusion}

This paper introduces \textsc{TAMIS}, a lightweight and interpretable anomaly detection system tailored for large-scale industrial time-series monitoring. Through extensive experiments, we addressed three core research questions: detector combination, feature-space design, and cross-domain robustness. More specifically, we demonstrate that \textsc{TAMIS}, effectively balances accuracy and efficiency, with the compact two-detector configuration achieving the best trade-off for scalable deployment, while maintaining strong performances in out-of-distribution conditions. Moreover, the proposed feature set, \textsc{TAMIS}$_\mathcal{F}$, outperforms existing alternatives such as \textsc{Catch22} and \textsc{TsFresh}, delivering a superior accuracy–throughput balance.
To encourage future research in this direction, we release an anonymized version of our benchmark. The latter is among the most extensive datasets of energy production time series for anomaly detection.
Finally, future work will focus on: (i) extending TAMIS to online learning; (ii) adapting TAMIS to multivariate time series.

\newpage

\bibliographystyle{IEEEtran}
\bibliography{biblio}


\end{document}